\documentclass[runningheads,a4paper]{llncs}
\usepackage[utf8]{inputenc}

\usepackage[pdftex,bookmarks,colorlinks=false]{hyperref}

\usepackage{amssymb} 
\usepackage{graphicx}

\usepackage{algpseudocode}

\usepackage{algorithmicx}
\usepackage{url}
\usepackage{float}

\urldef{\mails}\path|{m.lopuszynski,l.bolikowski}@icm.edu.pl|

\newcommand{\keywords}[1]{\par\addvspace\baselineskip
\noindent\keywordname\enspace\ignorespaces#1}

\newcommand{\strutDown}{\rule[-2ex]{0pt}{2ex}}
\newcommand{\strutUpDown}{\rule[-2ex]{0pt}{5ex}}
\begin{document}

\mainmatter

\title{Tagging Scientific Publications using Wikipedia and Natural
       Language Processing Tools.
       Comparison on the ArXiv Dataset}

\titlerunning{Tagging Scientific Publications using Wikipedia \dots}

\authorrunning{Tagging Scientific Publications using Wikipedia \dots}

\author{Micha{\l} {\L}opuszy\'nski \and \L{}ukasz Bolikowski}
\institute{
Interdisciplinary Centre for Mathematical and Computational Modelling,\\
University of Warsaw,\\
Pawi\'nskiego 5a, 02-106 Warsaw, Poland \\
\mails\\
\url{http://www.icm.edu.pl}}

\maketitle

\begin{abstract}
In this work, we compare two simple methods of tagging scientific
publications with labels reflecting their content.  As a first source of
labels Wikipedia is employed, second label set is constructed from the noun
phrases occurring in the analyzed corpus.  We examine the statistical
properties and the effectiveness of both approaches on the dataset
consisting of abstracts from 0.7 million of scientific documents deposited
in the ArXiv preprint collection. We believe that obtained tags can be
later on applied as useful document features in various machine learning
tasks (document similarity, clustering, topic modelling, etc.).


\bigskip
\begin{scriptsize}
This paper was published in
``Theory and Practice of Digital Libraries -- TPDL 2013 Selected Workshops'',
Communications in Computer and Information Science, Volume 416,
Springer 2014, pp 16-27. The final publication is available at
\href{http://dx.doi.org/10.1007/978-3-319-08425-1_3}{link.springer.com}.
\end{scriptsize}

\end{abstract}

\section{Introduction}
In this work, we present a study of two methods for contextualizing
scientific publications by tagging them with labels reflecting their
content.  First method is based on Wikipedia and second approach relies on
the noun phrases detected by the natural language processing (NLP) tools.
The motivation behind this study is threefold.

First, we would like to develop new meaningful features for document content
representation, which will go beyond basic bag of words approach.  The tags
can serve as such features, which later on can be employed for various
applications, e.g., determining document similarity, clustering, topic
modelling, and other machine learning tasks.  After the appropriate
filtering and ranking, obtained tags can also be used as keyphrases,
summarizing the document.

Our second goal is the comparison of the two approaches to tagging
publications with labels reflecting their content.  We employed two
methods, abbreviated hereafter NP and WIKI.  The NP approach relies on the
tags' dictionary generated from \emph{noun phrases} detected in analyzed
corpus using NLP tools. Similar approaches based on NLP techniques were
used , e.g., for keyphrase extraction~\cite{Barker2000,Hulth2003}.
Conversely, the WIKI method relies on readily available dictionary of
meaningful tags coming from filtered Wikipedia entries. Wikipedia was
already applied in many studies on conceptualizing and contextualizing
document collections. To name just a few recent examples, applications
include clustering~\cite{Spanakis2012,Spanakis2012a}, assigning readable
labels to the obtained document clusters~\cite{Nomoto2011,Nomoto2012},
facilitating classification~\cite{Wang2009}, or extracting
keywords~\cite{Joorabchi2013}.  However, not much is known about the
effectiveness of Wikipedia when it comes to processing scientific texts.
Especially, in the case of collections covering broad range of disciplines,
there is a lot of domain-specific vocabulary involved, usually beyond the
scope of interest of the average Internet user, i.e., Wikipedia reader and
author.  Example of such a broad collection is the ArXiv preprint
repository~\cite{arXiv}.  Manually creating the "gold standard" dictionary of
meaningful tags is a difficult task, as it would require a large team of
highly qualified experts from different disciplines. Therefore, we find that
it is insightful to compare the results obtained using the WIKI method with
the independent competitive NP approach.  Interesting questions include the
relative effectiveness of the WIKI/NP methods for different fields of
science, the average number of tags per document in both methods, the
typical tags missed by one of the methods and included in the other,
etc. Such a comparison can also show if the methods are complementary or if
one is superior than the other.

The third goal of this work is the analysis of statistical properties for
obtained tags. We look at distributions of number of different tags per
document. We also examine, if the Zipf's law is valid for the rank-frequency
curves of labels detected by both methods. It is also interesting to check,
if the aforementioned statistical properties are qualitatively similar for
the NP and WIKI tags.

The paper is organized as follows. In Sect.~\ref{sec:EmployedDatasets}, the
employed datasets are described. Afterwards, in
Sect.~\ref{sec:ProcessingMethods}, we provide the details of both tagging
procedures~---~the one based on Wikipedia (WIKI) and the complementary
approach based on the noun phrases (NP).  Comparison of both methods is the
subject of Sect.~\ref{sec:CompEffectiveness}.  Statistical properties of the
obtained tags are investigated in Sect.~\ref{sec:CompStatProperties}. The paper
is summarized in Sect.~\ref{sec:Summary}.

\section{Employed Datasets \label{sec:EmployedDatasets}}

\begin{figure}[!ht]
  \centering
  \includegraphics[width=0.70\linewidth]{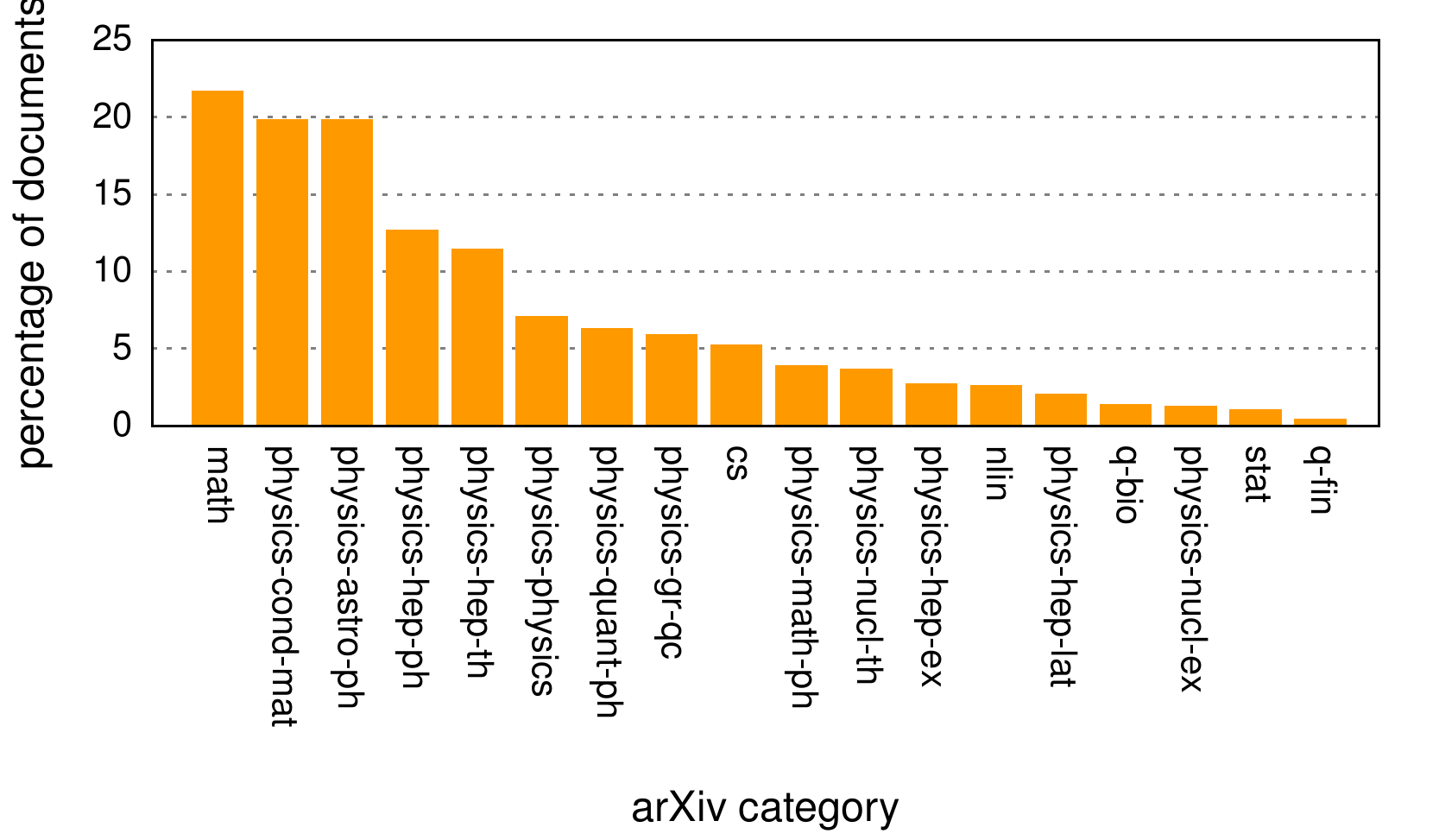}
  \caption{The percentage of documents marked with various ArXiv
           categories. Note that, since multiple categories per paper
           are possible, the sum of the numbers above exceeds 100\%. The
           labels for categories are explained in Table~\ref{tab:CatArXiv}.
           \label{fig:CatArXiv}}
\end{figure}

The ArXiv repository~\cite{arXiv} was started in 1991 by a physicist Paul
Ginsparg.  Originally, it was intended to host documents from the domain of
physics.  However, later on it gained popularity in other areas. Currently, it
hosts entries from physics, mathematics, computer science, quantitative
biology, quantitative finance, and statistics.  The content is not
peer-reviewed, however, many documents are simply preprints, published later
on in scientific journals or presented on conferences.  In this work, we
analyze the ArXiv publications metadata harvested via OAI/PMH protocol up to
the end of March 2012.  This made up to over 0.7 million of documents. For our
study, the distribution of the manuscripts across domains is of high
interest. For this purpose, we used \texttt{<setSpec>} field of the ArXiv XML
format, which gives a coarse-grained information about the field of document.
All the ArXiv coarse-grained categories together with their full-names are
presented in Table~\ref{tab:CatArXiv}.  The percentage of documents in each
category is displayed in Fig.~\ref{fig:CatArXiv}. The presented values do not
add up to 100\% since multiple categories per document are allowed.  In this
study, we have also employed Wikipedia. We have used raw data available from
the Wikipedia dump website, dated 2013.01.02.
\begin{table}[!ht]
\centering
\caption{The ArXiv categories and their abbreviations. \label{tab:CatArXiv}}
{\footnotesize
\begin{tabular}{cl}
\hline\noalign{\smallskip}
 Abbreviation  & Category Full Name \\
\hline\noalign{\smallskip}
    cs                 & Computer Science \\
    math               & Mathematics \\
    nlin               & Nonlinear Sciences \\
    \hspace{0.7cm}  physics-astro-ph  \hspace{0.7cm} & Astrophysics \\
    physics-cond-mat   & Condensed Matter Physics \\
    physics-gr-qc      & Physics --- General Relativity and Quantum Cosmology \\
    physics-hep-ex     & High energy Physics --- Experiment \\
    physics-hep-lat    & High energy Physics --- Lattice \\
    physics-hep-ph     & High energy Physics --- Phenomenology \\
    physics-hep-th     & High energy Physics --- Theory \\
    physics-math-ph    & Mathematical Physics \\
    physics-nucl-ex    & Nuclear Physics --- Experiment \\
    physics-nucl-th    & Nuclear Physics --- Theory \\
    physics-quant-ph   & Quantum Physics \\
    physics-physics    & Physics --- Other Fields \\
    q-bio              & Quantitative Biology \\
    q-fin              & Quantitative Finance \\
    stat               & Statistics \\
\hline
\end{tabular}}
\end{table}

\section{Processing Methods \label{sec:ProcessingMethods}}
Our processing methods consisted of three phases --- generating the
preliminary dictionary, cleaning the dictionary and tagging. Only the first
phase differentiated the two analyzed methods that is, the the approach
employing Wikipedia (WIKI) and the procedure making use of the
noun phrases (NP).

\begin{enumerate}
\item {\bf Generating the preliminary dictionary.} During this stage the
preliminary version of the dictionary used later on for labeling was obtained.
For the WIKI case, simply all multi-word entries form Wikipedia dump were
extracted. For the NP method all the abstract from ArXiv corpus were analyzed
using general purpose natural language processing library
OpenNLP~\cite{OpenNLPWebsite}, detecting all the noun phrases containing two
or more words. Noun phrases occurring in fewer than 4 documents were excluded
from the dictionary.

\item {\bf Cleaning the dictionary.} Clearly, on this level both dictionaries
contain a lot of non-informative entries.  Therefore, we apply a cleaning
procedure to both preliminary tag sets. For each tag we remove initial and
final words, if they belong to the set of stopwords. The labels which contain
only one word after such filtering are removed.  Then we use a simple
heuristic observation that good label candidates usually do not contain
stopword in the middle, see the study~\cite{Rose2010} for more details.  One
notable exception here is the word \emph{of}. We drop all entries according to
this heuristic rule.  Naturally, many far more sophisticated algorithms can be
employed here, e.g., matching grammatical pattern devised to select true
keywords, which could be employed, when the knowledge about the part-of-speech
classification is available~\cite{Justeson1995,Agrawal2012}.  However, the
simple stopword method worked well enough for us, especially that we are mostly
aiming at labels for further applications in machine learning and hence we can
afford having certain fraction of "bogus labels". The generated dictionaries
after the cleaning procedure contained around 5 million entries for the WIKI
method and 0.3 million for the NP case.

\item {\bf Tagging.} Finally we tag the analyzed corpus of ArXiv abstracts
with the obtained filtered dictionaries. In the process of tagging, we make use
of the Porter stemming~\cite{Porter1980}, to alleviate the problem of
different grammatical forms. All abstracts that contain sequence of words that
stems to the same roots as label contained in the WIKI/NP dictionary are tagged
with it.
\end{enumerate}

\section{Comparison of the WIKI and NP Tags Across Domains
         \label{sec:CompEffectiveness}}

\begin{figure}[!hb]
  \centering
  \includegraphics[width=0.70\linewidth]{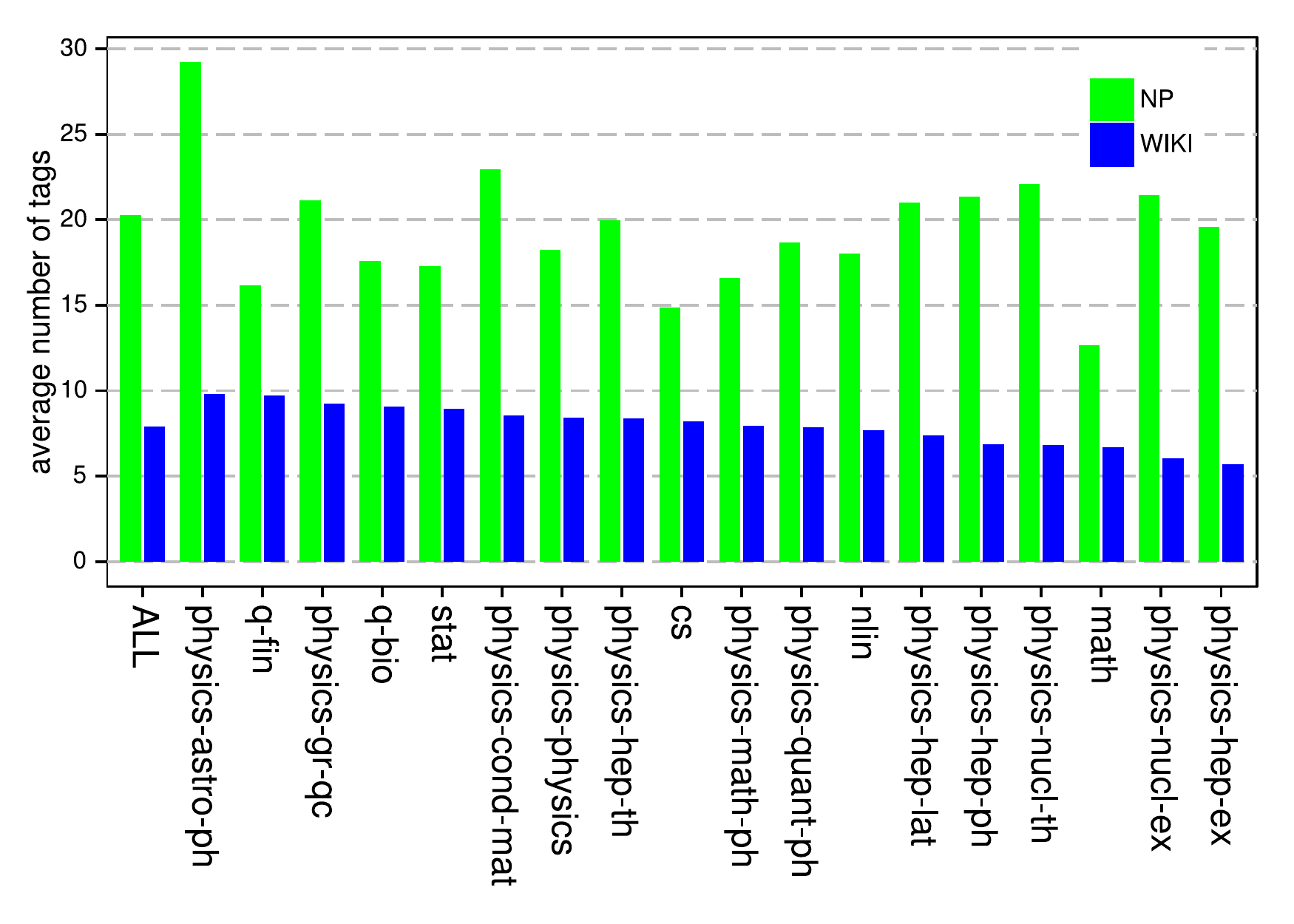}
  \caption{Average number of tags per article for the WIKI and NP cases
           separated into ArXiv categories. Note that categories are sorted
           according to the average number of WIKI tags in the descending order.
           \label{fig:AvgTags}}
\end{figure}

\begin{figure}[!ht]
  \centering
  \includegraphics[width=0.70\linewidth]{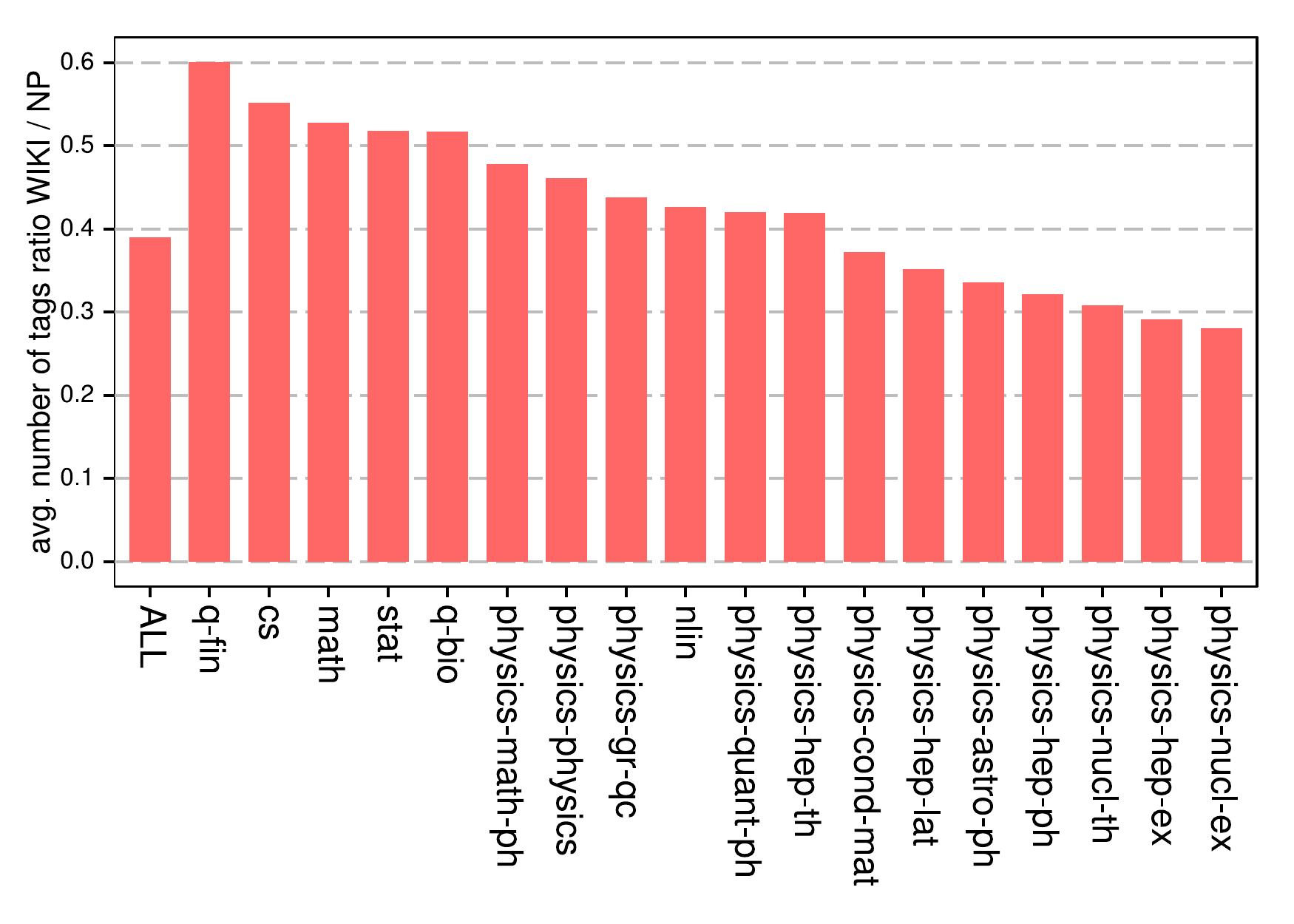}
  \caption{The ratio of the average number of the WIKI tags to the number of
           the NP tags for different ArXiv categories. The categories are
           sorted in the order of ascending ratio.  \label{fig:RatioAvgTags}}
\end{figure}

As a first step in the comparison of the WIKI and NP methods we calculated the
average number of tags per document. This quantity was examined across
different disciplines, the results are presented in Fig.~\ref{fig:AvgTags}.
The disciplines in Fig.~\ref{fig:AvgTags} are sorted according to the
average result for the WIKI method in ascending order.  This allows us to
observe that both methods are weakly correlated. In other words, if the WIKI
method gives high number of tags for certain category, it
does not imply that the NP approach yields high average as well.  This
observation can be quantified by calculating the correlation coefficient
between the average number of the WIKI and NP tags for each category, which
indeed turns out to have very low value of $\rho=0.13$.  Another conclusion
from Fig.~\ref{fig:AvgTags} is that clearly the NP method yields higher
number of tags across all the domains.  The average number of WIKI tags is
roughly in the range from 0.3 to 0.6 of the NP result. The exact ratios for
all the domains are visualized in Fig.~\ref{fig:RatioAvgTags}. The bar chart
is sorted according to the descending ratios. The sequence of disciplines
can be, to a certain extent, intuitively understood. The leading categories,
such as computer science and quantitative finance, are probably more
familiar to the average Internet user than experimental nuclear physics or
high-energy physics. Thus the coverage of the WIKI labels is also better in
these domains.  This indicates that various methods, relying on the
knowledge from Wikipedia and verified on the computer science texts (such
as, e.g., keyphrases in~\cite{Joorabchi2013}) can have considerably lower
performance when applied to documents from different scientific field.

\begin{table}[p]
\caption{Comparison of the top 10 most frequent tags in four categories.
The first column (Top WIKI) denotes labels occurring in the WIKI method.
The second column (Top NP) includes results produced by the NP method.  The
third column (Top WIKI-only) displays most frequent tags generated by the
WIKI method, but not by the NP. Finally, the fourth column shows the most
frequent NP results, not detected by the WIKI (Top NP-only). \label{tab:TopTags}}
{\footnotesize
\begin{tabular}{llll}
\hline\noalign{\smallskip}
Top WIKI  & Top NP & Top WIKI-only  & Top NP-only \\
\hline\noalign{\smallskip}
\multicolumn{4}{l}{{\bf cs} \strutUpDown } \\
lower bound              & lower bound               & state of art              & large scale               \\
upper bound              & upper bound               & degrees of freedom        & interference channel      \\
polynomial time          & polynomial time           & point of view             & time algorithm            \\
et al                    & et al                     & object oriented           & proposed algorithm        \\
sensor network           & sensor network            & quality of service        & proposed method           \\
logic programming        & logic programming         & order of magnitude        & hoc network               \\
wireless network         & wireless network          & game theory               & considered problem        \\
real time                & real time                 & Reed Solomon              & wireless sensor           \\
network coding           & network coding            & multi agent system        & channel state             \\
ad hoc \strutDown        & ad hoc                    & multi user                & capacity region           \\

\hline
\multicolumn{4}{l}{{\bf math} \strutUpDown } \\
Lie algebra              & Lie algebra               & Calabi Yau                & give rise                 \\
differential equation    & differential equation     & Navier Stokes             & higher order              \\
moduli space             & moduli space              & point of view             & initial data              \\
lower bound              & lower bound               & non negative              & infinitely many           \\
field theory             & field theory              & Cohen Macaulay            & new proof                 \\
finite dimensional       & finite dimensional        & algebraically closed      & over field                \\
sufficient condition     & sufficient condition      & degrees of freedom        & value problem             \\
upper bound              & upper bound               & self dual                 & large class               \\
Lie group                & Lie group                 & Gromov Witten             & time dependence           \\
two dimensional \strutDown   & two dimensional       & answered question         & mapping class             \\

\hline
\multicolumn{4}{l}{{\bf physics-nucl-ex}\strutUpDown } \\
cross section            & cross section             & equation of state         & heavy ion                 \\
Au Au                    & heavy ion                 & center of mass            & Au collisions             \\
heavy ion collision      & Au Au                     & order of magnitude        & ion collision             \\
form factor              & Au collisions             & degrees of freedom         & Au Au collision           \\
beta decay               & ion collision             & ultra relativistic        & transversal momentum      \\
elliptic flow            & Au Au collision           & Drell Yan                 & 200 GeV                   \\
high energies            & heavy ion collision       & time of flight            & relativistic heavy        \\
experimental data        & transversal momentum      & presented first           & relativistic heavy ions   \\
charged particle         & 200 GeV                   & long lived                & low energies              \\
nuclear matter \strutDown & form factor              & national laboratory       & Pb Pb                     \\

\hline
\end{tabular}
}
\end{table}

To further investigate the differences between the two methods we displayed
the most frequent tags generated by both methods in Table~\ref{tab:TopTags}.
In addition, we also included the most frequent tags generated uniquely by
each method, to be able to better judge the differences. We have performed
this analysis for three different ArXiv categories. We have selected
\texttt{cs} and \texttt{math} as they have high ratio of the WIKI/NP average
number of tags (we have neglected here \texttt{q-fin} since there is very
low number of documents from this field, see Fig.~\ref{fig:CatArXiv}). We
have also included \texttt{physics-nucl-ex}, as it is at the other end of the
spectrum, having very low aforementioned ratio of the WIKI/NP average number
of tags.  There are a couple of interesting observations, which can be made
from Table~\ref{tab:TopTags}. Note that Top WIKI and Top NP categories are
identical for \texttt{cs} and \texttt{math} categories, whereas for
\texttt{physics-nucl-ex} there are much different. In the latter case, the
top four WIKI tags occur also in the NP results, however, the NP adds a lot
of additional labels. They are mostly related to various kinds of nuclei
collision processes, which apparently are too specific to be described in
Wikipedia. Interestingly, the \emph{Au-Au} tag from the WIKI corresponds to
the article about one of the on-line auction portals and has nothing to do
with gold nuclei. Another interesting property is that the WIKI method is
much better at detecting surnames related to various theories, equations,
etc. In particular, this is visible for \texttt{math} and the WIKI-only
category, where four out of ten tags are related to surnames. Clearly, not
all of the above tags are perfect. It can be observed that noun-phrases
detector sometimes yields the fragments of actual noun phrase, e.g.,
\emph{hoc network} is a fragment of correct phrase \emph{ad hoc network},
\emph{time algorithm} comes from complexity statements, such as
\emph{polynomial time algorithm}, etc. There are also a few tags which do
not yield any information, e.g., \emph{et al}, \emph{point of view},
\emph{give rise}, \emph{initial data}, etc. If there is a need, their impact
can be reduced by improving the filtering procedure described in
Sect.~\ref{sec:ProcessingMethods}.

\begin{figure}[!b]
  \centering
  \includegraphics[width=0.49\linewidth]{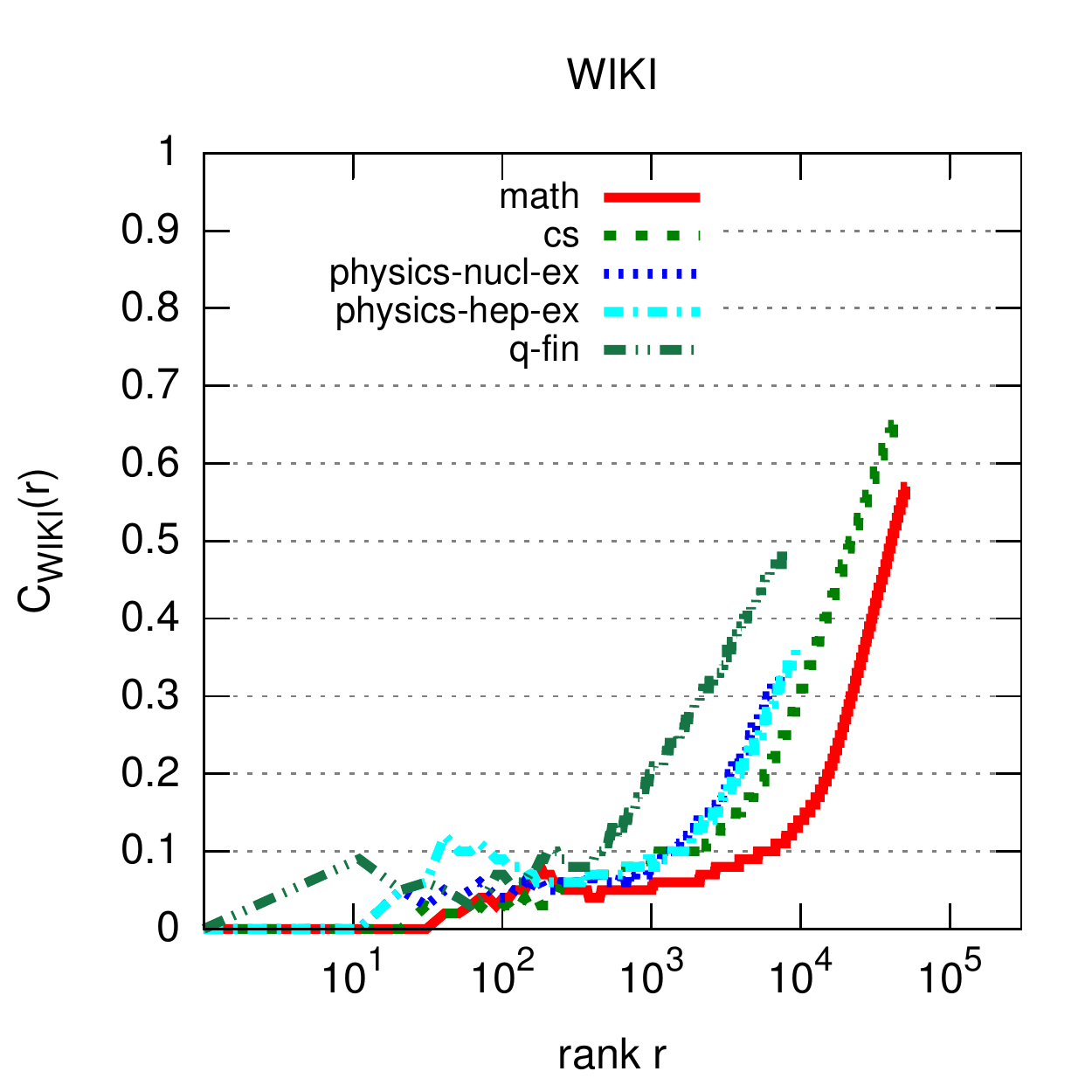}
  \includegraphics[width=0.49\linewidth]{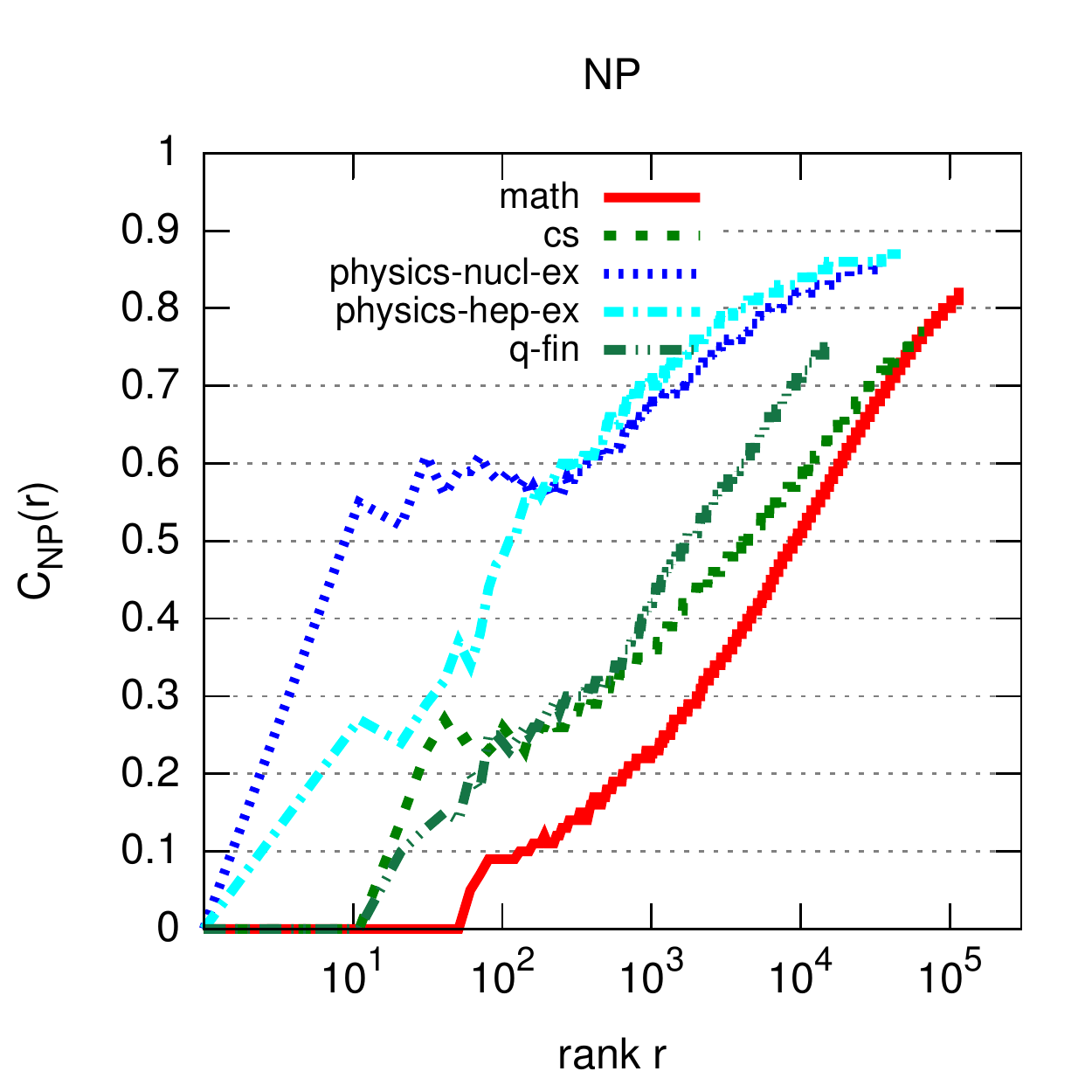}
  \caption{The dependence of $C_{\textsc{wiki}}$ (left panel) and
           $C_{\textsc{np}}$ (right panel) on rank $r$, i.e., the percentage of tags
           up to rank $r$ for the WIKI/NP method that were not detected by the
           other approach. Only a few sample categories were selected,
           including edge cases with the fastest and the slowest growing
           dependencies.  See Eq.~(\ref{eq:C}) and the main text for details
           \label{fig:C}.}
\end{figure}

As a final stage of the analysis we decided to address a question, to what
extent the tags generated by the WIKI and NP methods are different?
Table~\ref{tab:TopTags} suggests that in many categories top rank labels
might be similar. Larger deviations may get introduced for the less frequent
tags.  To examine this phenomenon, we propose the following measures that
describes the percentage of unique tags detected by each method up to rank
$r$
{\footnotesize
\begin{eqnarray}
  \label{eq:C}
  C_{\textsc{wiki}}(r)
      = \frac{\#(T_{\textsc{wiki}}(r) \setminus T_\textsc{np}(\infty))}{r},
  \hspace{0.5cm}
  C_{\textsc{np}}(r)
      = \frac{\#(T_{\textsc{np}}(r) \setminus T_\textsc{wiki}(\infty))}{r},
\end{eqnarray}}
\hspace{-0.3em}where $T_{\textsc{wiki}}(r)$ denotes the set of all tags up
to rank $r$ assigned by the WIKI method, $T_{\textsc{wiki}}(\infty)$ refers
to the set of all tags assigned by the WIKI method.  The meaning of
$T_{\textsc{np}}(r)$ and $T_{\textsc{np}}(\infty)$ is analogous, but refers
to the NP approach. The $C_{\textsc{wiki}}(r)$ function describes the
percentage of tags up to rank $r$, obtained from the WIKI method that were
not detected by the NP approach (independently of rank).  The
$C_\textsc{np}(r)$ has analogous meaning for the NP case.  The plots of the
above quantities for a few sample ArXiv categories are presented in
Fig.~\ref{fig:C}.
We have selected the categories in a way that the edge cases of the fastest
and the slowest growing dependencies are included.  The figures clearly show
that for the WIKI case the percentage of the unique tags is low, i.e. around
10\%, up to relatively high ranks, mostly $\sim 10^3$--$10^4$.  This
confirms the intuition that the relevant WIKI tags are indeed in majority
noun phrases. On the other hand, the curves for the NP case show a different
behaviour, the percentage of the unique tags grows much faster in this case,
indicating that they might yield much richer information. The 10\% level of
unique tags is exceeded for the ranks lower than $10^2$ for the most
categories.  However, to give the definitive statement about the quality of
the above tags, the domain experts should be consulted.

\section{Statistical Properties of the WIKI and NP
         Tags \label{sec:CompStatProperties}}
Tags can be expected to have similar statistical properties as ordinary words.
One of the universal properties observed for words is the so-called Zipf's
law, which states that the word frequency $f$ as a function of its rank $r$ in
the frequency table should exhibit power-law behaviour
\begin{equation}
\label{eq:ZipfsLaw}
f(r;A,N)= A\; r^{-N},
\end{equation}
where $A$ and $N$ are parameters. This type of simple dependency was observed
not only for words, but also keyphrases, e.g., in the PNAS
Journal bibliographic dataset~\cite{Zhang2008}. However, the detailed
investigation reveals that for large corpora, in particular when many
different authors and hence different styles are involved, the simple
model~(\ref{eq:ZipfsLaw}) might be insufficient to describe the frequency-rank
dependence throughout the whole $r$ variability range~\cite{Montemurro2001}.
Sometimes a few curves of the type~(\ref{eq:ZipfsLaw}) are necessary in order
to accurately describe the observed distribution throughout the whole rank
domain.

\begin{figure}[!ht]
  \centering
  \includegraphics[width=0.47\linewidth]{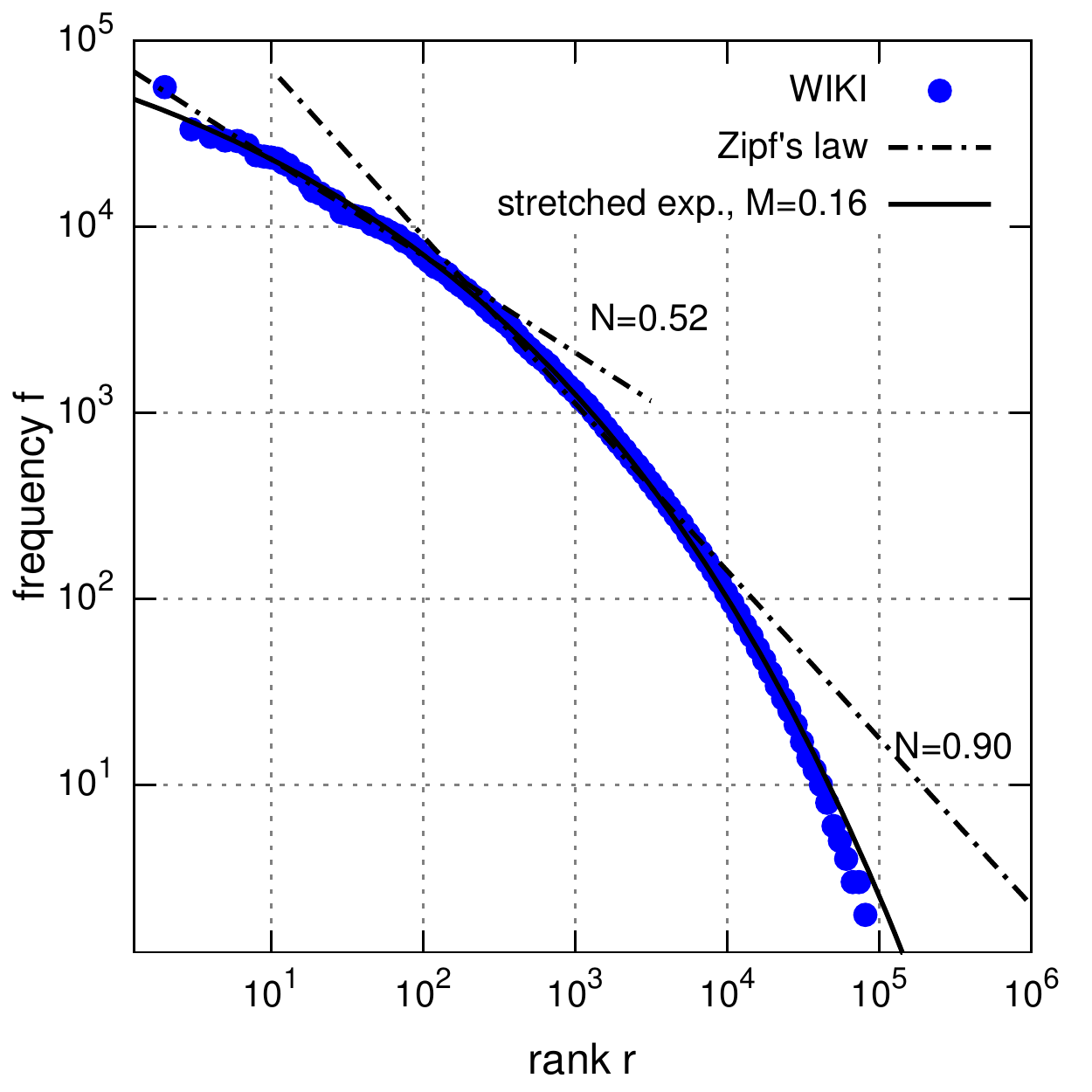}
  \includegraphics[width=0.47\linewidth]{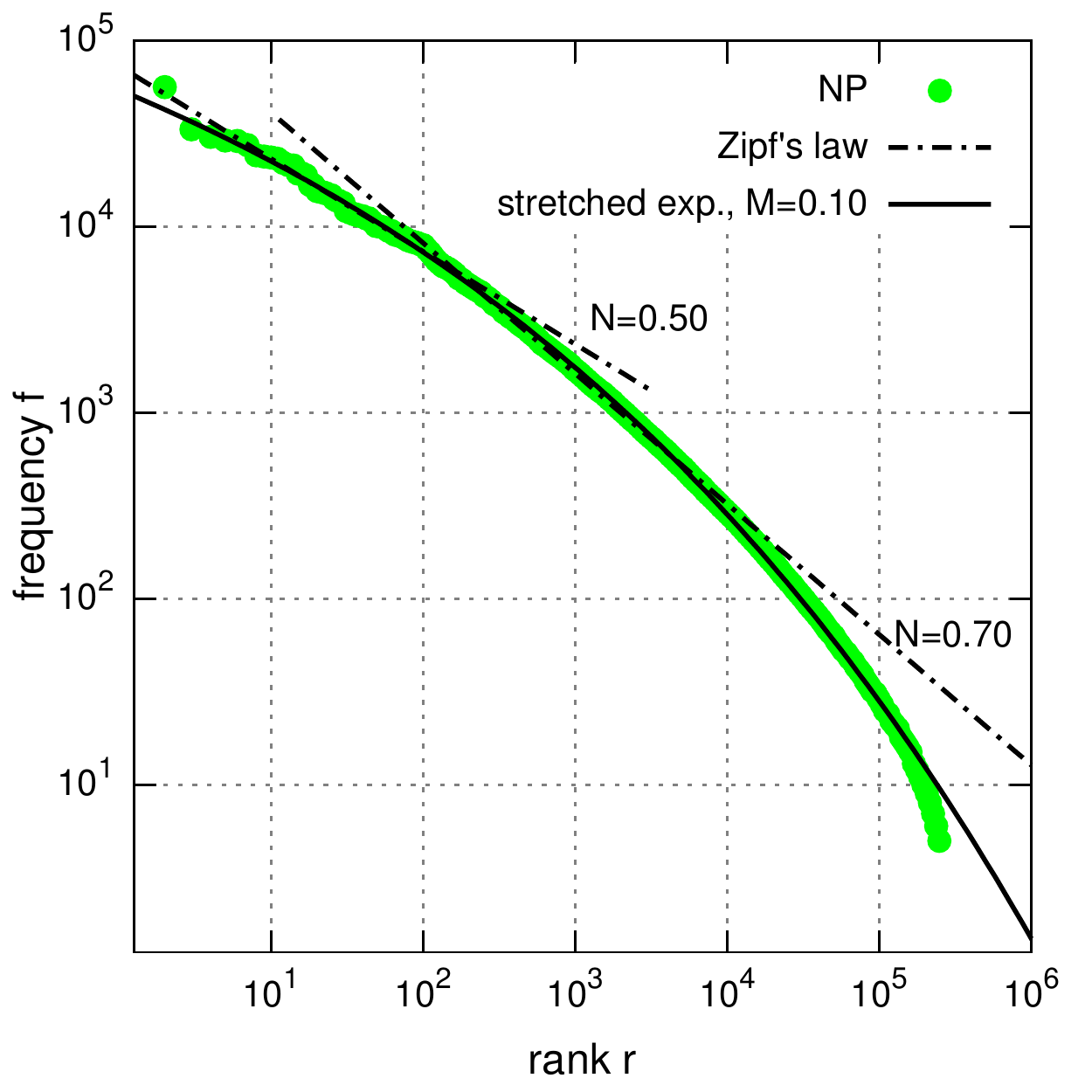}
  \caption{Comparison of the frequency dependence on rank
           observed for tags obtained from both approaches --- the WIKI (left panel)
           and the NP (right panel).
           The models fitted to the observed distributions are Zipf's law, see
           Eq.~(\ref{eq:ZipfsLaw}), and
           stretched exponential model, see Eq.~(\ref{eq:StretchedExponentialLaw}).
           \label{fig:CompZipf}}
\end{figure}

In the case of our tags, the observed rank-frequency dependencies are
presented in Fig.~\ref{fig:CompZipf}.
In both cases (WIKI and NP), the
crude approximation for the observed data was obtained using a combination
of two Zipf type curves for different rank regimes.  It turned out that up to
rank 100, the values of exponent $N$ were very similar in both cases and
approximately equal to $0.5$. However, for larger values the WIKI case showed
more rapid decay with $N=0.95$, as opposed to $N=0.73$ in the NP
case. Nevertheless, it is easily observed that a simple combination of the
Zipf type curves does not fit the data very well.  It turns out that the observed rank-frequency
dependencies are much better approximated by one of the alternatives to the
power-law~(\ref{eq:ZipfsLaw}), namely the stretched exponential distribution.
This type of distribution is used to describe large variety of phenomena from
physics to finance~\cite{Laherrere1998}. It was observed, e.g., for rank
distributions of radio/light emission intensities from galaxies, French and US
agglomeration sizes, daily Forex \mbox{US-Mark} price variation, etc.  The
stretched exponential model yields the following dependence of frequency on
rank
\begin{equation}
\label{eq:StretchedExponentialLaw}
f(r; C,D,M) = C \exp \left(- D\,r^M\right),
\end{equation}
with C, D, and M being parameters. As can be observed in
Fig.~\ref{fig:CompZipf}, this model fits the data much better. Similarly to the
Zipf's law, the value of the exponent for the NP case, which reads $M=0.12$,
is lower than for the WIKI, where $M=0.19$. This indicates slower decay
and "fatter tail" for the NP tags case.

\begin{figure}[!htb]
  \centering
  \includegraphics[width=0.45\linewidth]{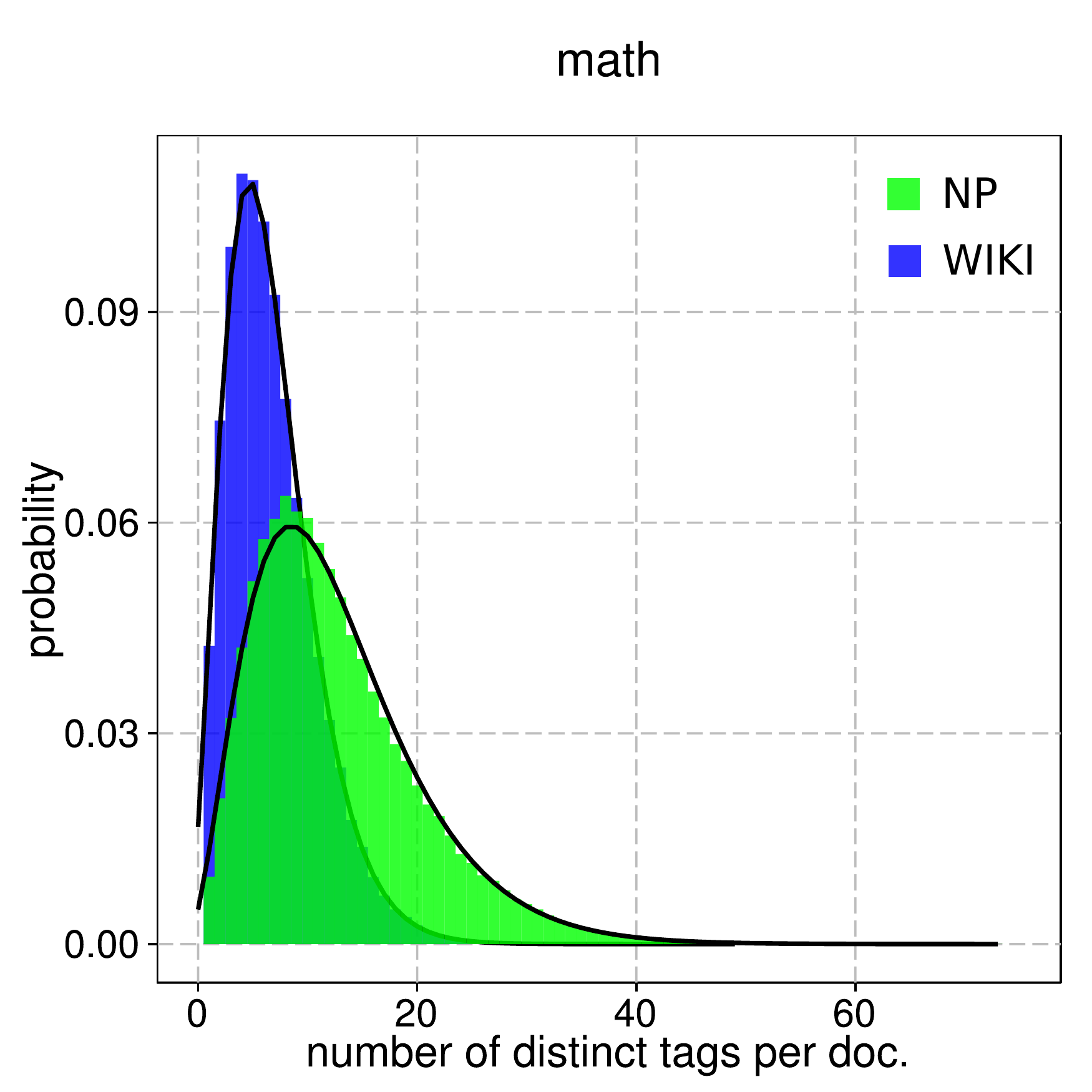}
  \includegraphics[width=0.45\linewidth]{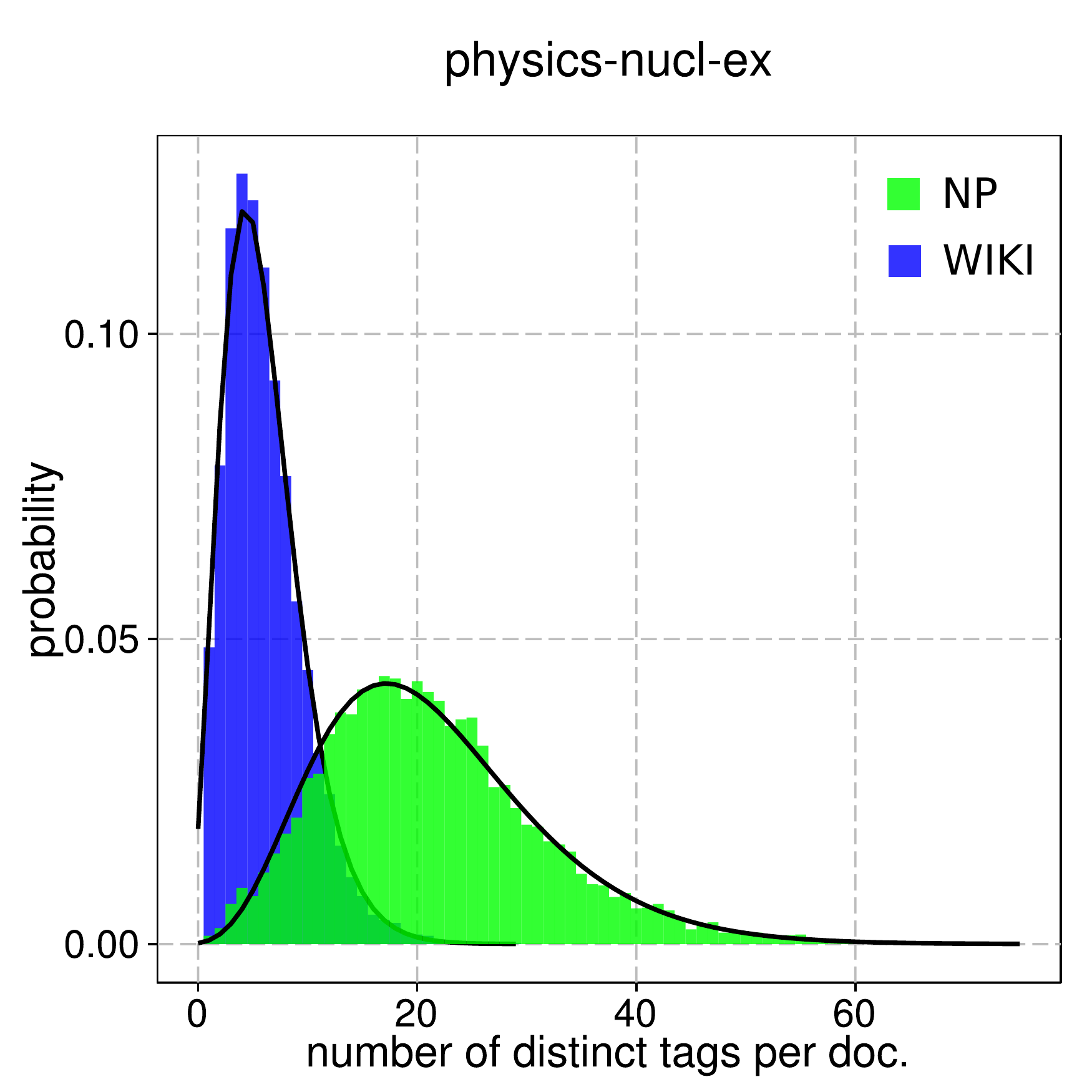}
  \caption{Distribution for the number of tags per document within two sample
           ArXiv categories \texttt{math} (left panel) and
           \texttt{physics-nucl-ex} (right panel). The distribution can be
           well approximated by the negative binomial distribution, see
           Eq.~(\ref{eq:NegativeBinomial}).  The black line represents the
           fits of this model to the observed data. \label{fig:NumTagsHist}}
\end{figure}

Another interesting statistical property of the generated tags is the
distribution for number of distinct labels per document. It turns out that,
even though the average tag counts per document are quite different for the
WIKI and NP methods (see Sect. \ref{sec:CompEffectiveness}), the
distributions in both cases come from the same family.  Observed histograms
can be approximated with the negative binomial distribution. According to this
model, the probability of finding document with $k$ tags reads
\begin{equation}
 \label{eq:NegativeBinomial}
 \textrm{Prob}(k;P,R)=\left ( {k + R - 1} \atop k \right ) P^R (1-P)^k,
\end{equation}
where $R>0$ and $P\in(0,1)$ are the parameters of the distribution. The
comparison of the above model with the observed histograms can be found
in Fig.~\ref{fig:NumTagsHist}.

\section{Summary and Outlook \label{sec:Summary}}
In this paper, we have compared two methods of tagging scientific
publications. First, abbreviated WIKI, was based on the multi-word entries
from Wikipedia. Second, referenced as NP, relied on the multi-word noun
phrases detected by the NLP tools.  We have focused on the effectiveness of
each method across domains and on the statistical properties of the obtained
labels.

When it comes to the effectiveness of the above methods, it turned out that
the NP approach yields higher average number of tags per document. The
difference is by a factor between two and three with respect to the WIKI
case. This strongly depends on domain. The WIKI tags coverage is better in the
areas more relevant to the Internet community, such as computer science or
quantitative finance than in more exotic domains such as nuclear experimental
physics. In addition, there is almost no correlation between the average
number of labels generated by the NP and WIKI methods, when separated  to
different scientific domains. This signals that results of both methods are to a
certain extent complementary.
When it comes to the differences in the obtained tags, it turns out that
high-rank labels from the WIKI method are usually also detected by the NP. The
notable, easy to understand exceptions are tags containing the complex of
surnames, such as \emph{Navier-Stokes}. Depending on the category, within the
first $10^3$~--~$10^4$ most frequent WIKI tags the percentage of the unique
labels is lower than 10\%. Conversely, for the NP method the number of unique
tags is much higher. Usually in the top 100 labels, there is already more than
10\% cases not found by the WIKI method.  However, the average level of "bogus
tags" seems higher for this method.  In particular, sometimes it yields broken
phrases such as \emph{hoc network} instead of \emph{ad hoc network}. The
development of more accurate filters for such cases or better part-of-speech
taggers/chunkers trained on scientific corpora could improve the method.

As far as the statistical properties are concerned, it turned out that both
the WIKI and NP methods exhibit qualitatively very similar behaviour.  The
dependence of the tag frequency on the tag rank can be approximated by the
Zipf's law, however, only in the limited rank range. To be able to cover the
whole rank domain the so-called stretched exponential model has to be
employed. It constitutes a good fit for both the WIKI and NP. Obtained curve
parameters indicate much slower decay ("fatter tail") for the NP method.
The investigation of the distribution for the number of tags per document
revealed that in both the WIKI and NP cases it follows quite closely
the negative-binomial model.

Overall, in our opinion, both the WIKI and NP methods seem useful, and to a
certain extent complementary.  In future we plan to apply the generated tags
as features, extending the simple bag of words document representation, in
various types of machine learning tasks (document similarity, clustering,
etc.). Verifying the performance on such tasks will enable for more definite
statement on the usefulness of both methods.

\noindent \sloppy {\bf Acknowledgement.} This research was carried out with the
 support of the "HPC Infrastructure for Grand Challenges of Science and
 Engineering (POWIEW)" Project, co-financed by the European Regional
 Development Fund under the Innovative Economy Operational Programme.
\bibliographystyle{unsrt}
\bibliography{biblio}

\begin{thebibliography}{10}

\bibitem{Barker2000}
Ken Barker and Nadia Cornacchia.
\newblock Using noun phrase heads to extract document keyphrases.
\newblock In HowardJ. Hamilton, editor, {\em Advances in Artificial
  Intelligence}, volume 1822 of {\em Lecture Notes in Computer Science},
  page~40. Springer Berlin Heidelberg, 2000.

\bibitem{Hulth2003}
Anette Hulth.
\newblock Improved automatic keyword extraction given more linguistic
  knowledge.
\newblock In {\em Proceedings of the 2003 conference on Empirical Methods in
  Natural Language Processing}, EMNLP '03, page 216, Stroudsburg, PA, USA,
  2003. Association for Computational Linguistics.

\bibitem{Spanakis2012}
Gerasimos Spanakis, Georgios Siolas, and Andreas Stafylopatis.
\newblock Exploiting {Wikipedia} knowledge for conceptual hierarchical
  clustering of documents.
\newblock {\em Comput. J.}, 55(3):299, March 2012.

\bibitem{Spanakis2012a}
Gerasimos Spanakis, Georgios Siolas, and Andreas Stafylopatis.
\newblock {DoSO}: a document self-organizer.
\newblock {\em J. Intell. Inf. Syst.}, 39(3):577, 2012.

\bibitem{Nomoto2011}
Tadashi Nomoto.
\newblock {WikiLabel}: an encyclopedic approach to labeling documents en masse.
\newblock In {\em Proceedings of the 20th ACM international conference on
  Information and knowledge management}, CIKM '11, page 2341, New York, NY,
  USA, 2011. ACM.

\bibitem{Nomoto2012}
Tadashi Nomoto and Noriko Kando.
\newblock Conceptualizing documents with {Wikipedia}.
\newblock In {\em Proceedings of the fifth workshop on Exploiting semantic
  annotations in information retrieval}, ESAIR '12, page~11, New York, NY, USA,
  2012. ACM.

\bibitem{Wang2009}
Pu~Wang, Jian Hu, Hua-Jun Zeng, and Zheng Chen.
\newblock Using {Wikipedia} knowledge to improve text classification.
\newblock {\em Knowledge and Information Systems}, 19(3):265, 2009.

\bibitem{Joorabchi2013}
Arash Joorabchi and Abdulhussain~E. Mahdi.
\newblock Automatic keyphrase annotation of scientific documents using
  {Wikipedia} and genetic algorithms.
\newblock {\em Journal of Information Science}, 39(3):410, 2013.

\bibitem{arXiv}
{arXiv} preprint server, {\texttt{http://arxiv.org}}.

\bibitem{OpenNLPWebsite}
Apache {OpenNLP}, {\texttt{http://opennlp.apache.org}}.

\bibitem{Rose2010}
Stuart Rose, Dave Engel, Nick Cramer, and Wendy Cowley.
\newblock {\em Automatic Keyword Extraction from Individual Documents}, page~1.
\newblock John Wiley and Sons, Ltd, 2010.

\bibitem{Justeson1995}
John~S. Justeson and Slava~M. Katz.
\newblock Technical terminology: some linguistic properties and an algorithm
  for identification in text.
\newblock {\em Natural Language Engineering}, 1(01):9, 2 1995.

\bibitem{Agrawal2012}
Rakesh Agrawal, Sreenivas Gollapudi, Anitha Kannan, and Krishnaram Kenthapadi.
\newblock Data mining for improving textbooks.
\newblock {\em SIGKDD Explor. Newsl.}, 13(2):7, May 2012.

\bibitem{Porter1980}
Martin Porter.
\newblock An algorithm for suffix stripping.
\newblock {\em Program: electronic library and information systems}, 14(3):130,
  1980.

\bibitem{Zhang2008}
Zi-Ke Zhang, Linyuan Lü, Jian-Guo Liu, and Tao Zhou.
\newblock Empirical analysis on a keyword-based semantic system.
\newblock {\em The European Physical Journal B}, 66(4):557, 2008.

\bibitem{Montemurro2001}
Marcelo~A. Montemurro.
\newblock Beyond the {Zipf}–{Mandelbrot} law in quantitative linguistics.
\newblock {\em Physica A: Statistical Mechanics and its Applications},
  300(3–4):567, 2001.

\bibitem{Laherrere1998}
J.~Laherrère and D.~Sornette.
\newblock Stretched exponential distributions in nature and economy: “fat
  tails” with characteristic scales.
\newblock {\em The European Physical Journal B}, 2(4):525, 1998.

\end{thebibliography}
\end{document}